# Factors Influencing User Willingness To Use SORA

*Gustave Florentin Nkoulou Mvondo*[*1] , Ben Niu[1]

**Abstract**

Sora promises to redefine the way visual content is created. Despite its numerous forecasted benefits, the drivers of user willingness to use the text-to-video (T2V) model are unknown. This study extends the extended unified theory of acceptance and use of technology (UTAUT2) with perceived realism and novelty value. Using a purposive sampling method, we collected data from 940 respondents in the US and analyzed the sample using covariance-based structural equation modeling and fuzzy set qualitative comparative analysis (fsQCA). The findings reveal that all hypothesized relationships are supported, with perceived realism emerging as the most influential driver, followed by novelty value. Moreover, fsQCA identifies five configurations leading to high and low willingness to use, and the model demonstrates high predictive validity, contributing to theory advancement. Our study provides valuable insights for developers and marketers, offering guidance for strategic decisions to promote the widespread adoption of T2V models.

**Keywords:** UTAUT2; Perceived realism; Novelty value; Willingness to use; SORA; Text-to-video model.

## 1. Introduction

In a captivating video clip, "A stylish woman in sunglasses confidently strides down a bustling Tokyo street, illuminated by the warm glow of neon lights and animated city signage. She is dressed in a black leather jacket, a long red dress, and black boots, and she carries a black purse. The street is damp and reflective, creating a mirror effect of the colorful lights. Numerous pedestrians are seen walking about" (Leffer, 2024). However, this description is not a scene from a television spot, a music video, or footage of any tangible event. The video clip was created by SORA, a text-to-video (T2V) model developed by OpenAI, the same company behind ChatGPT. SORA is capable of creating realistic, minute-long videos based on simple user prompts (Cotton & Crabtree, 2024). To train the model, OpenAI relied on a massive amount of videos and leveraged a generative pretrained transformer (GPT) to convert user prompts into detailed captions that are inputted into the model (OpenAI, 2024a). This enables it to generate high-quality, realistic videos that are hard to distinguish from those created by humans (Schomer, 2024). Just as ChatGPT has revolutionized communication dynamics (Niu & Mvondo, 2024), SORA is considered a game-changer in visual content creation.

---

[1]College of Management, Shenzhen University, China (*Corresponding Author: florent90@yahoo.fr)

SORA is not yet available to the general public but is already generating significant hype. Global search interest for the term "SORA" has surged to an impressive 100 as of February 2024 (Google Trends, 2024). The user community's interest is evident across various platforms such as LinkedIn, X, and YouTube, where discussions and content highlight its innovative and groundbreaking nature. However, despite the considerable enthusiasm and excitement surrounding SORA, there is limited understanding of the key factors influencing users' willingness to use the T2V model. In light of this, scholarly attention is required to explore the important drivers of user acceptance.

This study leverages the extended unified theory of acceptance and use of technology (UTAUT2) to explain users' willingness to use SORA. UTAUT2, developed by Venkatesh et al. (2013), aims to elucidate individuals' intentions to adopt technology. The model considers several hedonic, utilitarian, social, and individual factors as key determinants of an individual's intention to use technology. UTAUT2 is widely employed due to its generic nature and high predictive ability (Tamilmani et al., 2021). It has been validated in various technology acceptance contexts, such as research on autonomous vehicles, mobile apps (Tam et al., 2018), Metaverse (Arpaci et al., 2022), and virtual assistants (de Blanes Sebastián et al., 2022).

Several researchers have extended UTAUT2 with external and contextual factors to enhance its relevance in specific technology contexts (e.g., Arpaci et al., 2022; Shaw & Sergueeva, 2019). In this study, we expand the model by incorporating two factors that closely relate to the disruptive nature of SORA: perceived realism and novelty value. Perceived realism refers to the extent to which the visual elements of video content create a convincing and immersive depiction of reality. It encompasses factors such as the fidelity of visual representations, coherence of audiovisual elements, and overall believability of depicted scenes (Cho et al., 2014). On the other hand, novelty value measures the degree to which a technology is perceived as distinct from others due to its originality and freshness (Im et al., 2015; Wells et al., 2010).

This study aims to provide a comprehensive understanding of the factors influencing users' willingness to use SORA. It seeks to answer the question:

**RQ1: What key factors influence user willingness to use SORA?**

To address this question, this research develops an extended UTAUT2 model and tests it on a sample of 940 US respondents using covariance-based structural equation modeling (CB-SEM) and fuzzy set qualitative comparative analysis (fsQCA).

The originality and contributions of this research are as follows. Theoretically, as one of the pioneering studies exploring the adoption of T2V models, this research enriches our understanding of the determinants of users' willingness to use SORA. It highlights the positive impact of UTAUT2 factors, perceived realism, and novelty value on willingness to use. Notably, perceived realism emerged as a key factor, and we identified important configurations leading to high and low willingness to use SORA. Therefore, this research significantly contributes to the scholarly discourse on T2V models and human-AI interaction. From a managerial perspective, the study provides valuable insights for developers and marketers, guiding strategic decision-making to promote the widespread adoption of T2V models.

## 2. Literature review on text-to-video models

Kustudic and Mvondo (2024) define a T2V model as an AI model designed to generate video content from textual inputs. At its core, the model is engineered "to understand and simulate the physical world in motion, leveraging state-of-the-art AI techniques to produce high-quality videos that closely adhere to user prompts" (OpenAI, 2024). Although SORA is a notable advancement in T2V models (Kustudic & Mvondo, 2024), it should be noted that it is not the first of its kind. Earlier models such as "Emu by Meta, Gen-2 by Runway, Stable Video Diffusion by Stability AI, and recently Lumiere by Google, have paved the way in this domain" (Roth, 2024). Like generative pretrained transformer (GPT) models, "SORA utilizes a transformer architecture, enabling superior scaling performance and processing complex textual prompts" (OpenAI, 2024b). Moreover, SORA adopts a novel approach to representing videos and images, breaking them down into smaller units referred to as patches, akin to tokens in GPT (refer to Fig. 1). This unified representation of data "allows for training diffusion transformers on a broader range of visual data spanning different durations, resolutions, and aspect ratios" (Kustudic & Mvondo, 2024, p.3).

Research on T2V models is relatively scarce, primarily due to its recent emergence as a research topic. Most attention has been given to text-to-image models (Roth, 2024). Another reason is that advancements in T2V generation have been lagging compared to text-to-image generation (Singer et al., 2022; Wu et al., 2022). The existing research can be categorized into two main areas: T2V generation techniques and T2V applications and implications. In terms of T2V generation techniques, researchers have introduced new approaches to generate videos from text. For instance, Jiang et al. (2023) presented a technique they called Text2performer to generate vivid human video from text, while Wu et al. (2022) introduced the Tune-a-video technique. Moreover, Singer et al. (2022) introduced the Make-a-video approach, which is useful in accelerating the training of T2V models and generating videos that inherit "the vastness (diversity in aesthetic and fantastical depictions) of today's image generation models" (p. 1).

In terms of applications and implications, scholars have explored the opportunities, challenges, and implications of T2V models (e.g., Adetayo et al., 2024; Kustudic & Mvondo, 2024). For instance, Kustudic and Mvondo (2024) explain that T2V models, particularly SORA, offer various opportunities in "filmmaking, education, gaming, advertising, accessibility, healthcare, and social media content creation." Concerning risks, they acknowledge potential challenges, such as "misinformation, erosion of privacy, bias and discrimination, regulatory complexities, and dependence on technology" (p. 1)

As indicated by the literature review, research on T2V adoption is nonexistent. Specifically, (1) there is a lack of research providing a comprehensive understanding of the factors influencing user willingness to use SORA. This suggests that, although UTAUT2 has been extensively employed to explain user adoption of technology, (2) it has not been adapted in the context of T2V research. Additionally, (3) UTAUT2 has not yet been extended to include perceived realism and novelty value. The rationale for including these factors is that SORA can generate realistic videos that are challenging to distinguish from those created by humans (perceived realism) (Schomer, 2024). As such, the T2V model represents a disruptive and innovative technology that will revolutionize how video content is created (novelty value).

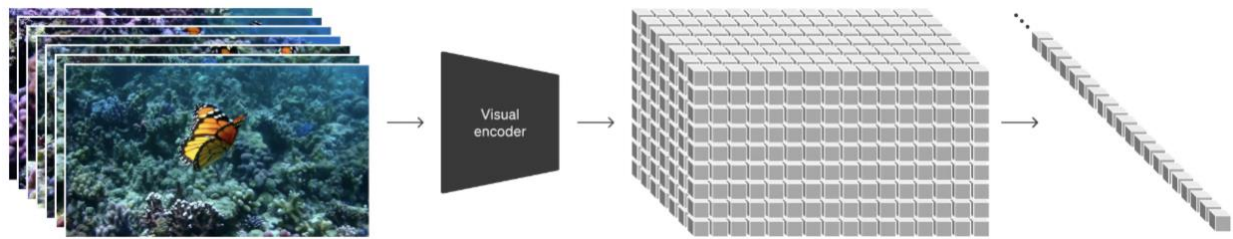

**Figure 1** Process of converting videos to lower dimensions and dividing them into patches
Source: SORA's technical report—OpenAI (2024b).

## 3. Theoretical foundation

3.1. Extended unified theory of acceptance and use of technology

Venkatesh et al. (2003) conducted a comprehensive analysis of the adoption theories, resulting in the formation of UTAUT, which is tailored to technology adoption in organizational contexts. The rise of consumer technologies necessitated a model adapted to this specific context (Venkatesh et al., 2012). Thus, the model was extended to include factors such as hedonic motivation, price value, and habit (Tamilmani et al., 2021). UTAUT2 has been employed in recent studies to explain user intention and behavior toward new technologies (e.g., Das & Datta, 2024; de Blanes Sebastián et al., 2022; Shareef et al., 2024; Suhail et al., 2024). For instance, Siyal et al. (2024) revealed that UTAUT2 factors and customization are critical determinants of the acceptance of mobile commerce applications, while de Blanes Sebastián et al., 2023 found that some UTAUT2 factors positively influence consumer adoption of mobile payment platforms.

This study adopts four factors from the UTAUT2 model: "performance expectancy, effort expectancy, social influence, and hedonic motivation." *Performance expectancy* "denotes the degree to which an individual believes that using a technology will help them to improve job performance goals." *Effort expectancy* "refers to how easily consumers expect to operate a particular technology." *Social influence* "represents the degree to which individuals perceive that important others believe they should use the new system." *Hedonic motivation* "reflects the fun or pleasure derived from using a technology" (Venkatesh et al., 2012, pp. 159-161).

## 4. Hypotheses development and conceptual model

4.1. Perceived realism

Perceived realism was "first investigated by communication scholars while studying the persuasive impact of narratives" (Daassi & Debbabi, 2021, p. 3). Subsequently, It has been applied to various technological contexts, including research on AR-based apps (Daassi & Debbabi, 2021), augmented reality (Orús et al., 2021), avatars (Pedram et al., 2020), and deepfake technology (Lu

& Chu, 2023). This research adapts the concept to the context of T2V and defines it as the extent to which the visual elements of video content create a convincing and immersive depiction of reality. Perceived realism comprises five dimensions: "plausibility, typicality, factuality, narrative consistency, and perceptual quality." This research specifically focuses on the perceptual quality aspect, as it "deals with the technical quality of media representation" (Lu & Chu, 2023, p.2). It evaluates the perceptual quality of the generated content, considering factors such as the fidelity of visual representations, coherence of audiovisual elements, and overall believability of depicted scenes (Cho et al., 2014).

The existing literature has not yet explored the impact of perceived realism on performance expectancy and willingness to use. However, we argue that when users perceive the video content generated by SORA as resembling real-world scenes and experiences, they are more likely to believe in the ability of the T2V model to perform well in meeting their video creation needs (performance expectancy). Additionally, as users perceive that SORA can effectively translate their textual prompts into compelling, lifelike video content, they are more likely to be willing to use the model. Based on this discussion, we hypothesize that:

**H1.** Perceived realism is positively related to performance expectancy.

**H2.** Perceived realism is positively related to user willingness to use SORA.

4.2. Novelty value

Novelty value is a new concept in information systems research. It is a fundamental aspect of any new technology (Wells, 2010). As mentioned earlier, novelty value refers to the extent to which users perceive technology as distinct from others due to its originality and freshness (Im et al., 2015; Wells et al., 2010). It represents a technological product's perceived uniqueness and freshness compared to existing alternatives. Pioneering psychological literature suggests that novelty elicits emotional responses (So et al., 2023; Wells et al., 2010). These emotional reactions arise from individuals perceiving a novel technology as an entirely unprecedented experience or a combination of previously encountered attributes presented in new ways (Wells et al., 2010).

The existing literature does not clearly explain the impact of novelty value on performance expectancy and user willingness to use technology. However, researchers explain that new technology, besides being novel, must address existing problems to be perceived as useful (McCarthy et al., 2018). SORA's novelty value lies in its ability to create minute-long videos based on simple prompts, which distinguishes it from traditional video creation methods. This innovative approach highlights its efficiency and flexibility, making it appealing to users. As users recognize SORA's unique capabilities, they are more likely to perceive it as a valuable asset that enhances their workflow, saves time, and expands creative possibilities. Based on this reasoning, we hypothesize that:

**H3.** Novelty value is positively related to performance expectancy.

Oyman et al. (2022) argue that consumers consistently seek novelty by exploring new product information or making new purchases. Furthermore, research on perceived novelty, a factor closely related to novelty value, indicates its influence on user intention to adopt technology (Fazal-e-Hasan et al., 2021). Therefore, we predict that the novelty value of SORA will impact user willingness to adopt the T2V model. Accordingly, we hypothesize that:

**H4.** Novelty value is positively related to user willingness to use SORA.

### 4.3. Performance expectancy

Performance expectancy is crucial in influencing the intention to adopt technologies (Venkatesh et al., 2003，2012).In the context of SORA, performance expectancy can be defined as the perceived benefits and practical value users anticipate from using the T2V model. Perceived usefulness is instrumental in shaping users' willingness to use SORA.When users believe that SORA can effectively fulfill their video creation aspirations and enhance their creative output by generating high-quality, realistic videos and visually captivating narratives, they are more likely to be willing to use The T2V model. Prior research has shown that performance expectancy influences user behavioral intentions (Camilleri, 2024; Siyal et al., 2024; Suhail et al., 2024). Accordingly, we propose that:

**H5.** Performance expectancy is positively related to user willingness to use SORA.

### 4.4. Effort expectancy

Effort expectancy is analogous to perceived ease of use (Venkatesh et al., 2003). It pertains to how user-friendly users perceive the technology to be. Effort expectancy is "crucial as it influences people's intention to use an innovation" (Senyo & Osabutey, 2020, p. 4). When applied to T2V models, usability involves users evaluating how easily they can create video content, navigate the interface, and interact with the model without encountering complexities or difficulties. Since SORA can generate content based on simple prompts, users are more likely to perceive it as user-friendly. This perception of effortless operation promotes user willingness to use by reducing feelings of frustration or stress typically associated with learning and using new technologies. Earlier studies have shown that effort expectancy influences user behavioral intentions (Alesanco-Llorente et al., 2023; Camilleri, 2024; Siyal et al., 2024). Accordingly, we hypothesize that:

**H6.** Effort expectancy is positively related to user willingness to use SORA.

### 4.5. Social influence

Social influence is "the degree to which individuals perceive that important others believe they should use the new system" (Venkatesh et al., 2003, p. 451). According to Suhail et al. (2024), "Individuals generally behave in a certain way to live up to the expectations of their peers, relatives, and family" (p. 4). They typically take into account the opinions of others when considering the adoption of new technologies. When users perceive positive feedback and endorsements from their friends, family, and influential figures on social media, they are more likely to become enthusiastic about using SORA. Additionally, engaging with others with similar interests or goals in content creation can create a sense of connection, leading to an increased desire to use the T2V model. Several scholars have emphasized the positive influence of social influence on users' behavioral intention to use technology (AlKheder et al., 2023; de Blanes Sebastián et al., 2023; Suhail et al., 2024). Thus, we propose the following:

**H7.** Social influence is positively related to user willingness to use SORA.

### 4.6. Hedonic motivation

Venkatesh et al. (2012) argue that hedonic motivation is crucial in technology adoption. Researchers have recognized the importance of engaging users hedonically (AlKheder et al., 2023;

Chakraborty et al., 2024; Mvondo et al., 2023a; Pop et al., 2023), as technology that solely provides utilitarian benefits may not sufficiently engage them for long-term usage. Users must derive fun and pleasure from using the technology (Siyal et al., 2024). Users are more likely to desire and use technology if they anticipate enjoyment and pleasure. If users envision using SORA as enjoyable, engaging, and entertaining, they may be motivated to use it. Similarly, if they perceive SORA as a means of creative expression, exploration, and experimentation, they may experience curiosity, fascination, and inspiration, which can increase their willingness to use the T2V model. Previous research has demonstrated that hedonic motivation influences user behavioral intentions (Basarir-Ozel et al., 2023; Siyal et al., 2024; Suhail et al., 2024). Accordingly, we advance that:

**H8.** Hedonic motivation is positively related to user willingness to use SORA (Fig. 2)

**Fig. 2** Extended UTAUT2 model

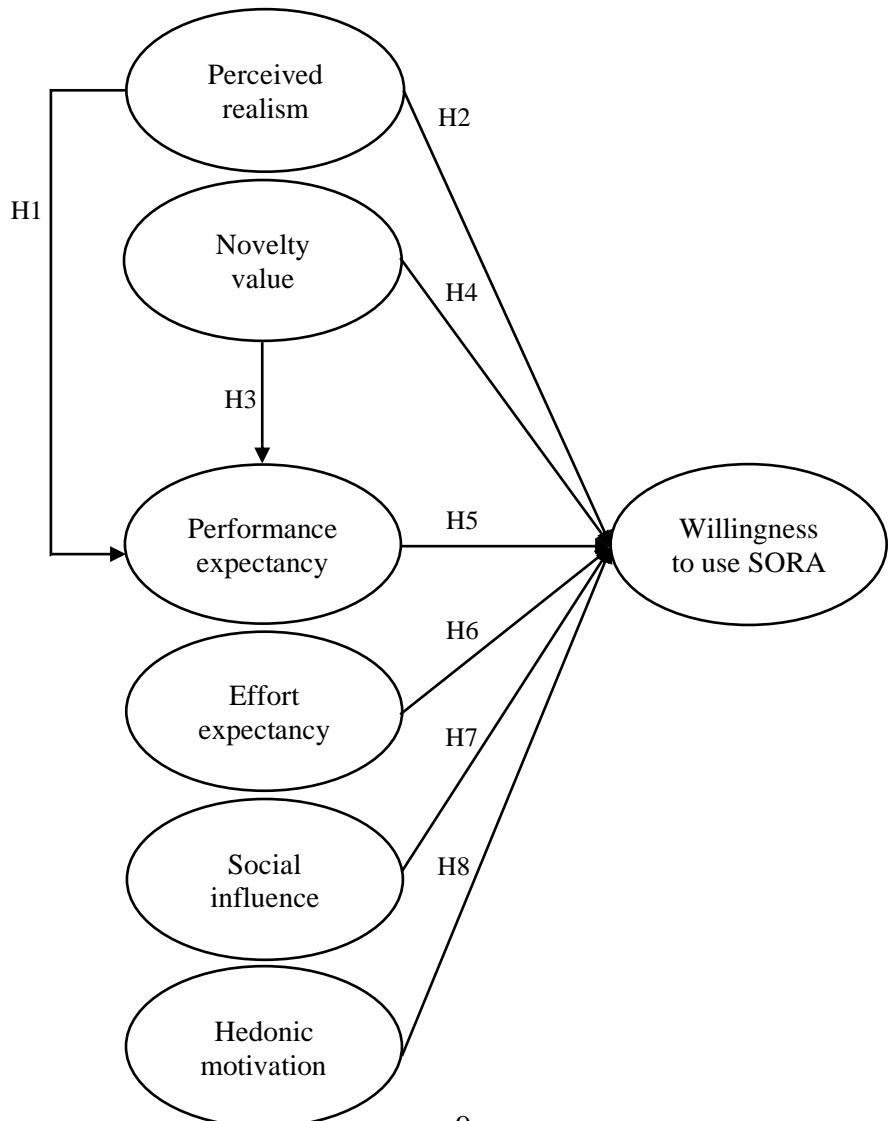

## 5. Methodology

### 5.1. Measures

We consulted existing literature to develop the research instrument and items, making minor adjustments for contextual relevance. We utilized a 7-point Likert-type scale to measure the items, ranging from "1 (strongly disagree)" to "7 (strongly agree)." To assess the UTAUT2 factors, we adapted Venkatesh et al.'s (2012) scale. Perceived realism was assessed with a five-item scale from Cho et al. (2012), while novelty value was measured with a four-item scale from Karjaluoto et al. (2019). Additionally, willingness to use was assessed using a three-item scale from Gursoy et al. (2019).

### 5.2. Pilot study

We initially carried out a pilot study with the following objectives: (1) assessing the validity of the measurement scales, (2) improving the quality of the questions, and (3) revising and adjusting the survey. The pilot study involved 95 respondents (46 females). Based on their feedback, we made further revisions to the questionnaire, ensuring that all constructs demonstrated a reliability scale exceeding 0.50, which was considered acceptable.

### 5.3. Data collection and sampling

Data were collected from US residents in April 2024 using a structured online questionnaire distributed via MTurk. We specifically targeted respondents who had prior knowledge of SORA by asking them if they had heard of SORA. Those who answered "No" were directed to the end of the survey.

To ensure the quality of our data, several measures were undertaken following Aguinis et al.'s (2021) guidelines. Firstly, an attention check was incorporated (an open-ended question asking respondents what the study was about). Secondly, we determined *a priori* that only respondents from the US, a native English-speaking nation, would be considered. Thirdly, we designed a concise questionnaire. Fourthly, scales with labeled "end" points were avoided. Lastly, respondents who passed the attention check and took between 165 seconds to 10 minutes to complete the survey were retained, aligning with screening practices to detect careless responses.

This meticulous process resulted in a total of 940 valid responses. The sample comprised 522 men (55.5%) and 418 women (44.5%). Within this cohort, 42.9% identified as white, while 29% identified as black or African American. Additionally, 78.3% fell within the age range of 18-40 years. The majority of respondents (87.1%) held at least a bachelor's degree and identified as students (35.6%), followed by employees (31.8%) and businesspeople (27.7%) (see Table 1).

**Table 1** Demographics

| Items | (N=940) | (%) |
|---|---|---|
| *Gender* | | |
| Male | 522 | 55.5 |
| Female | 418 | 44.5 |
| *Ethnicity* | | |
| White | 399 | 42.4 |
| Black or African American | 273 | 29 |
| Asian | 151 | 16.1 |
| Hispanic | 94 | 10 |
| Other | 23 | 2.4 |
| *Age* | | |
| 18-30 Years | 375 | 39.9 |
| 31-40 Years | 361 | 38.4 |
| 41-50 Years | 170 | 18.1 |
| Above 51 Years | 34 | 3.6 |
| 30 years or older | 19 | 2.9 |
| *Education* | | |
| High School or below | 101 | 10.7 |
| Bachelor's | 465 | 49.5 |
| Masters | 287 | 30.5 |
| Doctor (PhD) | 67 | 7.1 |
| Other | 20 | 2.1 |
| *Profession* | | |
| Student | 335 | 35.6 |
| Employee | 299 | 31.8 |
| Businesspeople | 260 | 27.7 |
| Unemployed | 20 | 2.1 |
| Other (e.g., Retired) | 26 | 2.8 |

5.4. Data analysis

This research employed a synergistic approach, combining CB-SEM and fsQCA for data analysis. This strategy has gained increasing popularity among management scientists aiming to address the inherent limitations of each method (e.g., Gandhi & Kar, 2024; Mvondo et al., 2023b), thereby enhancing the robustness of the research findings.

We used CB-SEM to explore causal relationships explaining variations in willingness to use. CB-SEM has limitations in identifying the critical combinations of factors that lead to high or low willingness to use. In contrast, fsQCA excels in identifying these combinations, standing out in research due to its ability to handle intricate relationships among multiple predictors and outcome variables. Unlike traditional symmetric analyses like regression and correlation, fsQCA considers multiple factors and combinations, providing a nuanced understanding of causality through fuzzy set theory and Boolean algebra (Ragin, 2009). This approach addresses limitations often encountered in correlation-based methods, such as overfitting, and offers a more comprehensive view of complex relationships.

Our data analysis unfolded in four stages: (1) assessing model reliability and validity, (2) testing hypothesized relationships, (3) conducting fsQCA, and (4) predictive validity. We used Amos version 29 for confirmatory factor analysis (CFA) and CB-SEM, SPSS version 27 for analyzing demographics and detecting common method bias (CMB), and fsQCA version 4.1 for qualitative comparative analysis.

## 6. Results

6.1. Common method bias

To ensure unbiased findings, we adopted a multi-step approach. First, we utilized professionally crafted surveys and refined the questionnaire by removing ambiguous or double-barreled questions to enhance respondents' comprehension and response accuracy. Additionally, we carried out a pilot test with potential respondents to optimize the survey design. Subsequently, we applied Harman's single-factor test, which indicated that the total variance explained by a single factor was only 10.092%, well below the widely accepted threshold of 50%. Furthermore, following Kock's (2015) recommendation, we conducted a comprehensive collinearity test, confirming that "all variance inflation factors were below the threshold of 3.3." Consequently, we confidently assert that there is no CMB concern in our research.

### 6.2. Measurement model

We evaluated the model fitness, construct reliability, and convergent and discriminant validity. The model demonstrated adequate fit indices, with the following values: chi-square ratio ($x2/df = 1.189$), incremental fit index (IFI = 0.997), comparative fit index (CFI = 0.997), Tucker Lewis index (TLI = 0.996), normed fit index (NFI = 0.978), standardized root mean square residual (SRMR = 0.018), and root mean square error of approximation (RMSEA = 0.014), all indicating satisfactory model fit. Moreover, the composite reliability (CR) values for the constructs exceeded 0.70, confirming the reliability of the measurement instruments. The outer loadings were above 0.704, which is considered acceptable. Each construct's average variance extracted (AVE) met the threshold (above 0.50), indicating convergent validity. Additionally, both the Fornell-Larcker criterion and the Heterotrait-Monotrait (HTMT) ratio supported the establishment of discriminant validity in the study (see Tables 2 and 3).

**Table 2** CFA

| Construct | Items | Loading | Alpha >0.7 | CR >0.7 | AVE >0.5 |
|---|---|---|---|---|---|
| PR | PR1 | 0.787 | 0.894 | 0.894 | 0.628 |
|  | PR2 | 0.795 |  |  |  |
|  | PR3 | 0.802 |  |  |  |
|  | PR4 | 0.805 |  |  |  |
|  | PR5 | 0.774 |  |  |  |
| NV | NV1 | 0.766 | 0.864 | 0.864 | 0.614 |
|  | NV2 | 0.788 |  |  |  |
|  | NV3 | 0.779 |  |  |  |
|  | NV4 | 0.801 |  |  |  |
| PE | PE1 | 0.784 | 0.840 | 0.840 | 0.637 |
|  | PE2 | 0.802 |  |  |  |
|  | PE3 | 0.808 |  |  |  |
| EE | EE1 | 0.805 | 0.848 | 0.848 | 0.650 |
|  | EE2 | 0.809 |  |  |  |
|  | EE3 | 0.805 |  |  |  |
| SI | SI1 | 0.802 | 0.832 | 0.832 | 0.623 |
|  | SI2 | 0.773 |  |  |  |
|  | SI3 | 0.792 |  |  |  |
| HM | HM1 | 0.814 | 0.842 | 0.842 | 0.640 |
| | HM2 | 0.790 | | | |
| | HM3 | 0.795 | | | |
| WU | WU1 | 0.798 | 0.846 | 0.846 | 0.647 |
| | WU2 | 0.805 | | | |
| | WU3 | 0.811 | | | |

**Table 3** Discriminant validity analysis

| Construct | 1 | 2 | 3 | 4 | 5 | 6 | 7 |
|---|---|---|---|---|---|---|---|
| 1. PR | **0.792** | | | | | | |
| 2. NV | 0.630 | **0.784** | | | | | |
| 3. PE | 0.541 | 0.556 | **0.798** | | | | |
| 4. EE | 0.576 | 0.561 | 0.511 | **0.806** | | | |
| 5. SI | 0.576 | 0.624 | 0.532 | 0.562 | **0.789** | | |
| 6. HM | 0.602 | 0.627 | 0.551 | 0.562 | 0.594 | **0.800** | |
| 7. WU | 0.630 | 0.573 | 0.530 | 0.534 | 0.554 | 0.564 | **0.805** |

6.3. Structural model

We first evaluated the predictive relevance of the model using the R-squared ($R^2$) approach. According to Mvondo et al. (2022), $R^2$ is "the proportion of an endogenous construct's variance explained by its predictor constructs in a regression model" (Mvondo et al., 2022a, 2022b). The analysis revealed that the extended UTAUT2 explains 58.9% of the variance in willingness to use SORA.

We tested our hypothesized relationships using the statistical bootstrap technique with a sample size of 5,000. Our findings support all hypothesized relationships. Specifically, we found that perceived realism (β=0.323), novelty value (β=0.101), performance expectancy (β=0.136), effort expectancy (β=0.102), social influence (β=0.127), and hedonic motivation (β=0.123) are positively related to user willingness to use SORA. Additionally, we found that perceived realism (β=0.328) and novelty value (β=0.404) are positively related to performance expectancy (see Table 4).

**Table 4** Hypotheses

| Hypothesized relationships | | Std. β | Std. Error | Supported |
|---|---|---|---|---|
| H1 | PR → PE | 0.328 | 0.041 | Yes |
| H2 | PR → WU | 0.323 | 0.044 | Yes |
| H3 | NV → PE | 0.404 | 0.040 | Yes |
| H4 | NV → WU | 0.101 | 0.039 | Yes |
| H5 | PE → WU | 0.136 | 0.027 | Yes |
| H6 | EE → WU | 0.102 | 0.026 | Yes |
| H7 | SI → WU | 0.127 | 0.034 | Yes |
| H8 | HM → WU | 0.123 | 0.036 | Yes |

6.4. fsQCA analysis

6.4.1. Calibration

Calibration is a prerequisite step before conducting fsQCA analysis. The data obtained from CB-SEM was transformed into fuzzy sets through calibration. We selected specific thresholds of "0.95, 0.50, and 0.05." These thresholds were chosen to "convert the data into a log-odds metric, ensuring that all values fall within the range of 0 to 1" (Pappas & Woodside, 2021). These thresholds are selected based on percentiles, enabling us to identify which values in our dataset correspond to the 0.95, 0.50 (median), and 0.05 percentiles. Specifically, we calculated the 95 %, 50 %, and 5 % of the measures and used these values as the three calibration thresholds (see Table 5).

**Table 5** Calibration

| | Full non-membership | Crossover point | Full membership |
|---|---|---|---|
| Calibration values at | 5% | 50% | 95% |
| PR | 2.00 | 5.40 | 6.20 |
| NV | 2.00 | 5.50 | 6.25 |
| PE | 1.67 | 5.33 | 6.33 |
| EE | 1.67 | 5.33 | 6.33 |
| SI | 2.00 | 5.33 | 6.33 |
| HM | 1.69 | 5.33 | 6.33 |
| WU | 1.67 | 5.33 | 6.33 |

6.4.2. Analysis of necessary conditions

The examination of necessary conditions aimed to identify if there are any specific causal conditions necessary for willingness to use. Following Ragin's (2009) criteria, a condition is necessary when its consistency exceeds 0.9. The findings in Table 6 show that no single condition qualifies as necessary for achieving willingness to use (WU). Additionally, no single condition is deemed necessary to negate willingness to use (~WU). These results suggest that no individual conditions can independently lead to WU or ~WU outcomes.

**Table 6** Analysis of necessary conditions

|  | Outcome variable: WU |  |  | Outcome variable: ~WU |  |
|---|---|---|---|---|---|
| Conditions | Consistency | Coverage | Conditions | Consistency | Coverage |
| PR | 0.762 | 0.749 | PR | 0.578 | 0.652 |
| ~PR | 0.646 | 0.572 | ~PR | 0.777 | 0.789 |
| NV | 0.773 | 0.734 | NV | 0.605 | 0.659 |
| ~NV | 0.641 | 0.586 | ~NV | 0.756 | 0.792 |
| PE | 0.750 | 0.748 | PE | 0.565 | 0.645 |
| ~PE | 0.644 | 0.564 | ~PE | 0.779 | 0.782 |
| EE | 0.746 | 0.743 | EE | 0.569 | 0.649 |
| ~EE | 0.648 | 0.567 | ~EE | 0.775 | 0.777 |
| SI | 0.748 | 0.738 | SI | 0.583 | 0.659 |
| ~SI | 0.655 | 0.578 | ~SI | 0.768 | 0.777 |
| HM | 0.730 | 0.746 | HM | 0.562 | 0.658 |
| ~HM | 0.666 | 0.570 | ~HM | 0.784 | 0.769 |

6.4.3. Results

Table 7 provides an overview of the findings, using simplified illustrations: "black circles denote the presence of a particular condition (●), crossed-out circles (⊗) signify its absence, and blank spaces represent a "do not care" situation, indicating the potential presence or absence of a causal condition" (Fiss, 2011). The table also includes information on each solution's raw consistency, similar to a regression coefficient, and coverage scores that reflect the magnitude of effects in hypothesis testing. Moreover, the overall coverage of the solution, analogous to the $R^2$ value in variable-based methodologies, helps assess the influence of identified configurations on user acceptance.

The findings reveal three configurations associated with high willingness to use and two configurations linked to low willingness, meeting "acceptable consistency (>0.8) and coverage (>0.2) levels" (Rasoolimanesh et al. (2021). In combinations that lead to a high willingness to use, novelty value, effort expectancy, and hedonic motivation consistently emerge across all three configurations, while perceived realism and social influence are present in two. Additionally, combination 3 emerges as the most favored and effective solution. This configuration represents a group of people who are willing to use SORA due to the combined presence of "***perceived realism***

*×novelty value ×performance expectancy ×effort expectancy ×social influence ×and hedonic motivation."* This configuration achieves a consistency score of 0.941 and a raw coverage of 0.43, indicating that 43% of respondents favor this strategy.

For configurations resulting in low willingness to use, the results indicate that the absence of any factors negatively impacts user willingness to use SORA.

**Table 7** Sufficient configurations for high and low user willingness

| Configurations | Solutions for high WU<br>Model: WU = f (PR, NV, PE, EE, SI, HM)<br>Frequency cut-off: 5<br>Consistency cut-off: 0.85 | | | Solutions for Low WU<br>Model: ~ WU = f (NV, PE, EE, SI, HM, PA)<br>Frequency cut-off: 9<br>Consistency cut-off: 0.85 | |
|---|---|---|---|---|---|
| | 1 | 2 | 3 | 1 | 2 |
| PR | ● | ⊗ | ● | ⊗ | ⊗ |
| NV | ● | ● | ● | ⊗ | ⊗ |
| PE | ⊗ | ⊗ | ● | ⊗ | ● |
| EE | ● | ● | ● | | ⊗ |
| SI | ⊗ | ● | ● | ⊗ | ● |
| HM | ● | ● | ● | ⊗ | ⊗ |
| Raw coverage | 0.248 | 0.242 | 0.430 | 0.496 | 0.313 |
| Unique coverage | 0.033 | 0.032 | 0.234 | 0.233 | 0.050 |
| Consistency | 0.864 | 0.853 | 0.941 | 0.930 | 0.941 |
| solution coverage | 0.521 | | | 0.546 | |
| solution consistency | 0.887 | | | 0.927 | |

6.5. Predictive validity of the model

The analysis involves developing a model that effectively fit our dataset and indicates predictive validity. Predictive validity assesses "how well the model predicts outcomes when other data samples are applied to a similar model" (Gandhi & Kar, 2024, p. 15). We used fsQCA 4.1 to test the predictive validity. Following Pappas & Woodside's (2021) guidelines, we first randomly split the dataset into "two samples at a ratio of 70:30, creating a subsample and a holdout sample" (Gandhi & Kar, 2024). The causal configurations derived from the subsample were subsequently tested on the holdout sample. Next, after obtaining the findings, we tested the predictive validity. According to Woodside (2014), models with a consistency exceeding 0.80 are valuable and contribute to advancing theory. An illustration of consistent outcomes is depicted in the plot shown in Fig. 3, where the model exhibits a "consistency of 0.934448 and a coverage of 0.430571." This indicates that the model aligns with 93.44% of the data and explains 43.05% of the total memberships related to willingness to use.

**Fig 3** Illustration of the outcomes

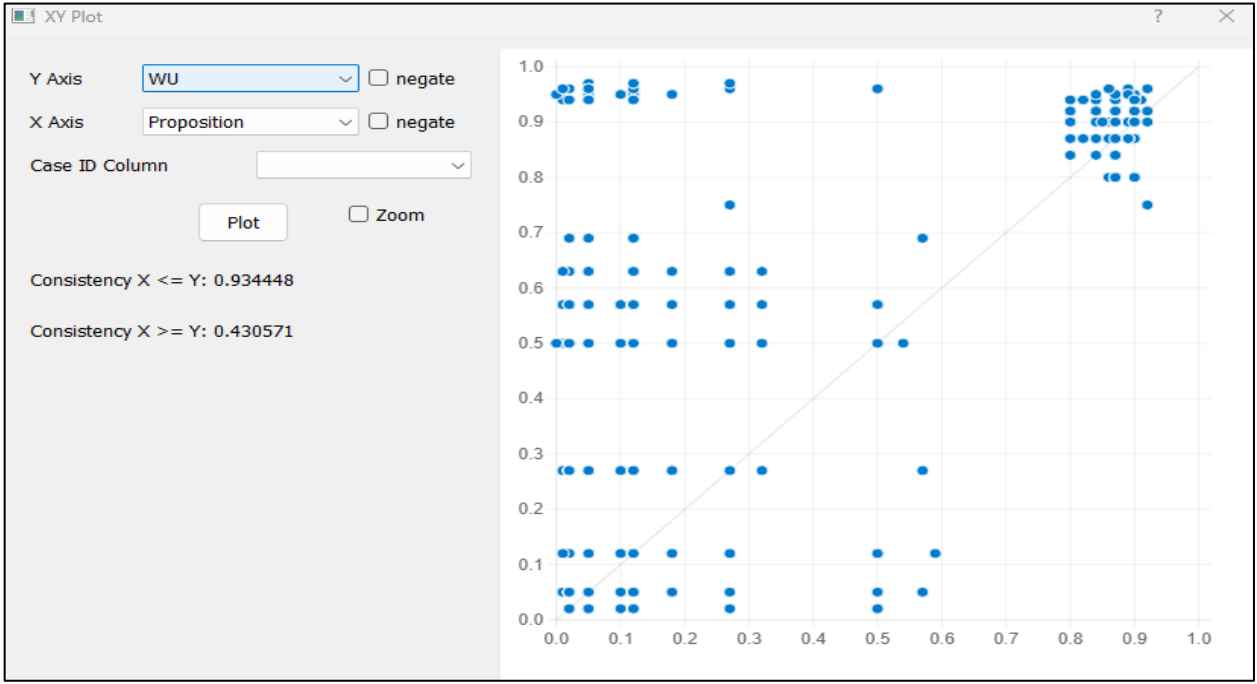

## 7. Discussion

This paper explored the important drivers of user willingness to use SORA by extending UTAUT2 with perceived realism and novelty value. The findings from CB-SEM demonstrated that perceived realism is a critical driver of performance expectancy (H1) and willingness to use (H2). This suggests that users perceive the T2V model as useful and are willing to use it when they find its generated content realistic and engaging. Similarly, novelty value, emphasizing the perceived uniqueness and freshness of SORA's capabilities, emerged as an essential driver of performance expectancy (H3) and willingness to use (H4). These findings emphasize the critical role of perceived innovation and distinctiveness in driving user acceptance of novel technologies like SORA.

Moreover, this research reaffirmed the significance of established UTAUT2 factors—performance expectancy (H5), effort expectancy (H6), social influence (H7), and hedonic motivation (H8)—in influencing behavioral intentions. Users' positive expectations regarding the practical benefits (performance expectancy) and user-friendliness (effort expectancy) of SORA, along with social influences and hedonic motivations, were identified as predictors of willingness to use the T2V model. These findings align with prior research on technology acceptance (e.g., Siyal et al., 2024; Suhail et al., 2024) and highlight the holistic nature of user perceptions and motivations in shaping adoption behaviors.

Furthermore, the findings from fsQCA revealed five complex configurations of factors associated with high and low willingness to use SORA. Notably, in configurations leading to high willingness, the third solution, which includes all the identified factors, emerged as the best-performing strategy, and the negation of any of these factors led to low willingness to use. This underscores the paramount importance of our extended UTAUT2 model in explaining user willingness to use the T2V model. This importance was further confirmed through the predictive validity test, as the model demonstrated high predictive validity, thus making it valuable in theory advancement (Woodside, 2014).

## 8. Implications and future research lines

### 8.1. Theoretical implications

From a research standpoint, this study makes a pioneering contribution by identifying perceived realism as a predictor of performance expectancy and willingness to use. Notably, this factor emerged as the most influential driver of willingness to use. Initially explored in communication studies to explain narrative persuasion (Daassi & Debbabi, 2021), perceived realism has been relatively underexplored in technology adoption. Earlier research has focused on its application in assessing the realism of deepfake content (Lu & Chu, 2023) and user attitudes toward augmented reality (Daassi & Debbabi, 2021). However, its role in driving user adoption and perceptions of technology usability has not been thoroughly examined. This research is the first to demonstrate the significant impact of perceived realism on performance expectancy and user willingness to use SORA, thereby making substantial contributions to the scholarly discourse on T2V models and human-AI interaction. These findings extend our understanding of how realistic visual representations shape individuals' acceptance of novel technologies like SORA. Furthermore, these observations could enable the development of mechanisms to effectively integrate perceived realism into marketing strategies aimed at fostering the widespread adoption of T2V models.

This research also identifies novelty value as a crucial predictor of performance expectancy and willingness to use SORA. Therefore, it advances the extant literature by adding these two factors to the list of outcomes of novelty value. Novelty value has received less attention, even though prior research has recognized its importance in emphasizing the uniqueness and freshness of new technology features (Im et al., 2015; So et al., 2023; Wells et al., 2010). Thus, our findings provide a novel perspective that enhances our understanding of the unique context of T2V models and establishes a solid foundation for future research endeavors in this domain.

Furthermore, this research pioneers the examination of user adoption of T2V models. Previous studies have predominantly focused on developing T2V generation techniques (e.g., Jiang et al., 2023; Singer et al., 2022; Wu et al., 2022) and exploring their applications and implications (e.g., Kustudic & Mvondo, 2024). Thus, this research makes an important contribution to the discourse on T2V models and paves the way for further research in this domain.

This research offers a unique extension of UTAUT2 specifically tailored to the context of T2V models, providing evidence of its relevance in explaining user adoption of T2V models. This expands the list of relevant theoretical frameworks that can elucidate the adoption of T2V models. Our extended UTAUT2 model accounts for 58.9% of the variance in willingness to use SORA. This suggests that while our model provides valuable insights into the predictors of willingness to use, further investigation is needed to enhance its explanatory power.

Lastly, from a methodological standpoint, the article goes beyond mere descriptive, textual, or correlation analyses. Instead, it employs a multifaceted approach, integrating CB-SEM and fsQCA. It conducts a new analysis to test "the predictive validity of the model through fsQCA" (Pappas & Woodside, 2021). By utilizing these advanced analytical techniques, this study delves deeper into the complexities of T2V model adoption, yielding richer insights and nuanced understandings of this phenomenon. Consequently, this methodological approach contributes additional knowledge and establishes a robust foundation for future investigations in this field.

8.2. Managerial implications

The present research provides actionable insights for developers and marketers. The findings underscore the notable impact of identified factors—perceived realism, novelty value, performance expectancy, effort expectancy, social influence, and hedonic motivation—on user willingness to use. Therefore, attention should be given to these factors, with particular emphasis on perceived realism, which emerged as the most critical driver of willingness to use SORA. To capitalize on this, developers should prioritize enhancing the realism of generated content by improving visual fidelity, audiovisual coherence, and overall believability. Investing in technologies that optimize these aspects will bolster users' confidence in the T2V model's ability to effectively meet their video creation needs. Novelty value is also of utmost importance, as it consistently appeared across the three configurations leading to a high willingness to use SORA. To leverage this aspect, developers should continuously innovate through advanced research in "machine learning, natural language processing, and computer vision" to enhance accuracy and creativity. Additionally, they should focus on scalability, performance optimization, and intuitive user interfaces (UI) to ensure seamless integration and usability.

For marketers, the focus should be on leveraging the enhanced perceived realism and novelty value of T2V models in promotional efforts. Emphasizing these aspects in marketing campaigns and user communications can pique curiosity and interest among potential adopters. Marketers should highlight how T2V technology like SORA can revolutionize content creation processes, offering users a unique and immersive way to bring their ideas to life. This promotional effort will position T2V models as valuable tools that empower users to create engaging video content effortlessly, driving greater acceptance and adoption and ultimately contributing to their success in the market.

Moreover, companies developing T2V models should prioritize the solutions identified in this study, placing particular emphasis on the last strategy, which underscores the importance of all the identified factors. This configuration emerged as the most efficient and effective approach in driving a high willingness to use SORA. Additionally, insights gleaned from configurations associated with low willingness to use also underscore the pivotal role of all these factors, as the absence of any of them can lead to reduced willingness. By focusing on these configurations, companies can promote the widespread adoption of T2V models.

## 8.3. Future research lines

This research has several limitations that suggest avenues for future research endeavors. One limitation of the study is its focus on user willingness, which represents intention rather than actual use behavior. Future research could examine actual use behavior to provide a more comprehensive understanding of adoption patterns and factors influencing sustained usage over time. Another limitation is that our extended UTAUT2 model explains 58.9% of the variance in user willingness to use SORA. Further research is needed to enhance its explanatory power by considering additional factors that may influence technology adoption. Finally, we focused solely on SORA, while there are other T2V models, such as Lumiere and Emu. Future research could investigate the intention of users to adopt these models.

**Reference**

Adetayo, A. J., Enamudu, A. I., Lawal, F. M., & Odunewu, A. O. (2024). From text to video with AI: the rise and potential of Sora in education and libraries. *Library Hi Tech News*. https://doi.org/10.1108/LHTN-02-2024-0028

Aguinis, H., Villamor, I., & Ramani, R. S. (2021). MTurk research: Review and recommendations. *Journal of Management*, *47*(4), 823–837. https://doi.org/10.1177/0149206320969787

Alesanco-Llorente, M., Reinares-Lara, E., Pelegrín-Borondo, J., & Olarte-Pascual, C. (2023). Mobile-assisted showrooming behavior and the (r) evolution of retail: The moderating effect of gender on the adoption of mobile augmented reality. *Technological Forecasting and Social Change*, *191*, 122514. https://doi.org/10.1016/j.techfore.2023.122514

AlKheder, S., Bash, A., Al Baghli, Z., Al Hubaini, R., & Al Kader, A. (2023). Customer perception and acceptance of autonomous delivery vehicles in the State of Kuwait during COVID-19. *Technological Forecasting and Social Change*, *191*, 122485. https://doi.org/10.1016/j.techfore.2023.122485

Arpaci, I., Karatas, K., Kusci, I., & Al-Emran, M. (2022). Understanding the social sustainability of the Metaverse by integrating UTAUT2 and big five personality traits: A hybrid SEM-ANN approach. *Technology in Society*, *71*, 102120. https://doi.org/10.1016/j.techsoc.2022.102120

Basarir-Ozel, B., Nasir, V. A., & Turker, H. B. (2023). Determinants of smart home adoption and differences across technology readiness segments. *Technological Forecasting and Social Change*, *197*, 122924. https://doi.org/10.1016/j.techfore.2023.122924

Camilleri, M. A. (2024). Factors affecting performance expectancy and intentions to use ChatGPT: Using SmartPLS to advance an information technology acceptance framework.

*Technological Forecasting and Social Change*, *201*, 123247. https://doi.org/10.1016/j.techfore.2024.123247

Chakraborty, D., Polisetty, A., & Rana, N. P. (2024). Consumers' continuance intention towards metaverse-based virtual stores: A multi-study perspective. *Technological Forecasting and Social Change*, *203*, 123405. https://doi.org/10.1016/j.techfore.2024.123405

Cho, H., Shen, L., & Wilson, K. (2014). Perceived realism: Dimensions and roles in narrative persuasion. *Communication Research*, *41*(6), 828–851. https://doi.org/10.1177/0093650212450585

Cotton, R., & Crabtree, M. (2024). "What is AI's Sora? How it works, use cases, alternatives & more." Available at: (https://www.datacamp.com/blog/openai-announces-sora-text-to-video-generative-ai-is-about-to-go-mainstream) (Last accessed, April 23, 2024)

Daassi, M., & Debbabi, S. (2021). Intention to reuse AR-based apps: The combined role of the sense of immersion, product presence and perceived realism. *Information & Management*, *58*(4), 103453. https://doi.org/10.1016/j.im.2021.103453

Das, S., & Datta, B. (2024). Application of UTAUT2 on Adopting Artificial Intelligence Powered Lead Management System (AI-LMS) in passenger car sales. *Technological Forecasting and Social Change*, *201*, 123241. https://doi.org/10.1016/j.techfore.2024.123241

de Blanes Sebastián, M. G., Antonovica, A., & Guede, J. R. S. (2023). What are the leading factors for using Spanish peer-to-peer mobile payment platform Bizum? The applied analysis of the UTAUT2 model. *Technological Forecasting and Social Change*, *187*, 122235. https://doi.org/10.1016/j.techfore.2022.122235

Fazal-e-Hasan, S. M., Amrollahi, A., Mortimer, G., Adapa, S., & Balaji, M. S. (2021). A multi-method approach to examining consumer intentions to use smart retail technology. *Computers in Human Behavior*, *117*, 106622. https://doi.org/10.1016/j.chb.2020.106622

Fiss, P. C. (2011). Building better causal theories: A fuzzy set approach to typologies in organization research. *Academy of Management Journal*, *54*(2), 393–420. https://doi.org/10.5465/amj.2011.60263120

Gandhi, M., & Kar, A. K. (2024). Dress to impress and serve well to prevail–Modelling regressive discontinuance for social networking sites. *International Journal of Information Management*, *76*, 102756. https://doi.org/10.1016/j.ijinfomgt.2024.102756

Google Trends. (2024). Term "SORA." Available at (https://trends.google.com/trends/explore?q=sora&hl=en) (Last accessed, April 23, 2024)

Gursoy, D., Chi, O. H., Lu, L., & Nunkoo, R. (2019). Consumers acceptance of artificially intelligent (AI) device use in service delivery. *International Journal of Information Management*, *49*, 157–169. https://doi.org/10.1016/j.ijinfomgt.2019.03.008

Im, S., Bhat, S., & Lee, Y. (2015). Consumer perceptions of product creativity, coolness, value and attitude. *Journal of Business Research*, *68*(1), 166–172. https://doi.org/10.1016/j.jbusres.2014.03.014

Jiang, Y., Yang, S., Koh, T. L., Wu, W., Loy, C. C., & Liu, Z. (2023). Text2performer: Text-driven human video generation. *Proceedings of the IEEE/CVF International Conference on Computer Vision*, 22747–22757.

Karjaluoto, H., Shaikh, A. A., Saarijärvi, H., & Saraniemi, S. (2019). How perceived value drives the use of mobile financial services apps. *International Journal of Information Management*, *47*, 252–261. https://doi.org/10.1016/j.ijinfomgt.2018.08.014

Kock, N. (2015). Common method bias in PLS-SEM: A full collinearity assessment approach. *International Journal of E-Collaboration (Ijec)*, *11*(4), 1–10. https//doi.org10.4018/ijec.2015100101

Kustudic, M., & Mvondo, G. F. N. (2024). A Hero Or A Killer? Overview Of Opportunities, Challenges, And Implications Of Text-To-Video Model SORA. *Authorea Preprints*. 10.36227/techrxiv.171207528.88283144/v2

Leffer, L (2024). "Everything to Know About OpenAI's New Text-to-Video Generator, Sora." Available at: (https://www.scientificamerican.com/article/sora-openai-text-video-generator/) (Last accessed, April 23, 2024)

Lu, H., & Chu, H. (2023). Let the dead talk: How deepfake resurrection narratives influence audience response in prosocial contexts. *Computers in Human Behavior*, *145*, 107761. https://doi.org/10.1016/j.chb.2023.107761\

McCarthy, M., Chen, C. C., & McNamee, R. C. (2018). Novelty and usefulness trade-off: Cultural cognitive differences and creative idea evaluation. *Journal of Cross-Cultural Psychology*, *49*(2), 171–198. https://doi.org/10.1177/0022022116680479

Mvondo, G. F. N, Jing, F., Hussain, K., & Raza, M. A. (2022). Converting tourists into evangelists: Exploring the role of tourists' participation in value co-creation in enhancing brand evangelism, empowerment, and commitment. *Journal of Hospitality and Tourism Management*, *52*, 1–12. https://doi.org/10.1016/j.jhtm.2022.05.015

Mvondo, G. F. N., Jing, F., & Hussain, K. (2023a). What's in the box? Investigating the benefits and risks of the blind box selling strategy. *Journal of Retailing and Consumer Services*, *71*, 103189. https://doi.org/10.1016/j.jretconser.2022.103189

Mvondo, G.F.N. Niu, Ben and Eivazinezhad, Salman. (2023b). Exploring The Ethical Use of LLM Chatbots In Higher Education. *Available at SSRN4548263 (2023)*. http://dx.doi.org/10.2139/ssrn.4548263

Niu, B., & Mvondo, G. F. N. (2024). I Am ChatGPT, the ultimate AI Chatbot! Investigating the determinants of users' loyalty and ethical usage concerns of ChatGPT. *Journal of Retailing and Consumer Services*, *76*, 103562. https://doi.org/10.1016/j.jretconser.2023.103562

OpenAI. (2024). "Creating video from text." Available at (https://openai.com/sora) (Last accessed, April 20, 2024)

Orús, C., Ibáñez-Sánchez, S., & Flavián, C. (2021). Enhancing the customer experience with virtual and augmented reality: The impact of content and device type. *International Journal of Hospitality Management*, *98*, 103019. https://doi.org/10.1016/j.ijhm.2021.103019

Oyman, M., Bal, D., & Ozer, S. (2022). Extending the technology acceptance model to explain how perceived augmented reality affects consumers' perceptions. *Computers in Human Behavior*, *128*, 107127. https://doi.org/10.1016/j.chb.2021.107127

Pappas, I. O., & Woodside, A. G. (2021). Fuzzy-set Qualitative Comparative Analysis (fsQCA): Guidelines for research practice in Information Systems and marketing. *International Journal of Information Management*, *58*, 102310. https://doi.org/10.1016/j.ijinfomgt.2021.102310

Pedram, S., Palmisano, S., Perez, P., Mursic, R., & Farrelly, M. (2020). Examining the potential of virtual reality to deliver remote rehabilitation. *Computers in Human Behavior*, *105*, 106223. https://doi.org/10.1016/j.chb.2019.106223

Pop, R.-A., Hlédik, E., & Dabija, D.-C. (2023). Predicting consumers' purchase intention through fast fashion mobile apps: The mediating role of attitude and the moderating role of COVID-19. *Technological Forecasting and Social Change*, *186*, 122111. https://doi.org/10.1016/j.techfore.2022.122111

Ragin, C. C. (2009). *Redesigning social inquiry: Fuzzy sets and beyond*. University of Chicago Press.

Rasoolimanesh, S. M., Ringle, C. M., Sarstedt, M., & Olya, H. (2021). The combined use of symmetric and asymmetric approaches: Partial least squares-structural equation modeling and fuzzy-set qualitative comparative analysis. *International Journal of Contemporary Hospitality Management*, *33*(5), 1571–1592. https://doi.org/10.1108/IJCHM-10-2020-1164

Roth, E. (2024) "OpenAI introduces Sora, its text-to-video model" available at: (https://www.theverge.com/2024/2/15/24074151/openai-sora-text-to-video-ai) (Last accessed, April 21, 2024)

Schomer, A. (2024). "Sora ai videos easily confused with real footage in survey test (exclusive)" Available at (https://variety.com/vip/sora-ai-video-confusion-human-test-survey-1235933647/) (Last accessed, April 1, 2024)

Shareef, M. A., Das, R., Ahmed, J. U., Mishra, A., Sultana, I., Rahman, M. Z., & Mukerji, B. (2024). Mandatory adoption of technology: Can UTAUT2 model capture managers behavioral intention? *Technological Forecasting and Social Change*, *200*, 123087. https://doi.org/10.1016/j.techfore.2023.123087

Shaw, N., & Sergueeva, K. (2019). The non-monetary benefits of mobile commerce: Extending UTAUT2 with perceived value. *International Journal of Information Management*, *45*, 44–55. https://doi.org/10.1016/j.ijinfomgt.2018.10.024

Singer, U., Polyak, A., Hayes, T., Yin, X., An, J., Zhang, S., Hu, Q., Yang, H., Ashual, O., & Gafni, O. (2022). Make-a-video: Text-to-video generation without text-video data. *ArXiv Preprint ArXiv:2209.14792*.

Siyal, A. W., Chen, H., Shah, S. J., Shahzad, F., & Bano, S. (2024). Customization at a glance: Investigating consumer experiences in mobile commerce applications. *Journal of Retailing and Consumer Services*, *76*, 103602. https://doi.org/10.1016/j.jretconser.2023.103602

So, J., Shim, M., & Song, H. (2023). Diffusion of COVID-19 misinformation: Mechanisms for threat-and efficacy-related misinformation diffusion. *Computers in Human Behavior*, *149*, 107967. https://doi.org/10.1016/j.chb.2023.107967

Suhail, F., Adel, M., Al-Emran, M., & AlQudah, A. A. (2024). Are students ready for robots in higher education? Examining the adoption of robots by integrating UTAUT2 and TTF using a hybrid SEM-ANN approach. *Technology in Society*, *77*, 102524. https://doi.org/10.1016/j.techsoc.2024.102524

Tam, C., Santos, D., & Oliveira, T. (2020). Exploring the influential factors of continuance intention to use mobile Apps: Extending the expectation confirmation model. *Information Systems Frontiers*, *22*, 243–257. https://doi.org/10.1007/s10796-018-9864-5

Tamilmani, K., Rana, N. P., Wamba, S. F., & Dwivedi, R. (2021). The extended Unified Theory of Acceptance and Use of Technology (UTAUT2): A systematic literature review and theory evaluation. *International Journal of Information Management*, *57*, 102269. https://doi.org/10.1016/j.ijinfomgt.2020.102269

Venkatesh, V., Morris, M. G., Davis, G. B., & Davis, F. D. (2003). User acceptance of information technology: Toward a unified view. *MIS Quarterly*, 425–478. https://doi.org/10.2307/30036540

Venkatesh, V., Thong, J. Y. L., & Xu, X. (2012). Consumer acceptance and use of information technology: extending the unified theory of acceptance and use of technology. *MIS Quarterly*, 157–178. https://doi.org/10.2307/41410412

Wells, J. D., Campbell, D. E., Valacich, J. S., & Featherman, M. (2010). The effect of perceived novelty on the adoption of information technology innovations: a risk/reward perspective. *Decision Sciences*, *41*(4), 813–843. https://doi.org/10.1111/j.1540-5915.2010.00292.x

Williams, S. (2024). "Unveiling the Dangers of OpenAI's SORA" Available at: (https://medium.com/@StephanyWilliamsWrites/unveiling-the-dangers-of-openais-sora-0866c1f5a53d) (Last accessed, April 1, 2024)

Woodside, A. G. (2014). Embrace• perform• model: Complexity theory, contrarian case analysis, and multiple realities. *Journal of Business Research*, *67*(12), 2495–2503. https://doi.org/10.1016/j.jbusres.2014.07.006

Wu, J. Z., Ge, Y., Wang, X., Lei, S. W., Gu, Y., Shi, Y., Hsu, W., Shan, Y., Qie, X., & Shou, M. Z. (2023). Tune-a-video: One-shot tuning of image diffusion models for text-to-video generation. *Proceedings of the IEEE/CVF International Conference on Computer Vision*, 7623–7633.